\documentclass[letterpaper, 10 pt, conference]{ieeeconf}  % Comment this line out if you need a4paper

\IEEEoverridecommandlockouts                              % This command is only needed if 
\usepackage{xcolor}
\usepackage{siunitx}
\usepackage{booktabs}
\usepackage{mathtools}
\usepackage{amssymb}

\usepackage{cite}
\usepackage{stix}
\usepackage{graphicx}
\usepackage{tabularx,booktabs}

\title{\LARGE \bf
Improving Convolutional Neural Networks for Fault Diagnosis by Assimilating Global Features
}

\author{Saif S. S. Al-Wahaibi$^{1}$ and Qiugang Lu$^{1, \dagger}$ % <-this % stops a space
\thanks{*This work was supported by the Texas Tech University}% <-this % stops a space
\thanks{$^{1}$S. Al-Wahaibi and Q. Lu are with the Department of Chemical Engineering, Texas Tech University, Lubbock, TX 79409, USA.
        {\tt\small Email: Saif.Al-Wahaibi@ttu.edu; Jay.Lu@ttu.edu}}
    \thanks{$^{\dagger}$Corresponding author: Q. Lu}
}

\begin{document}

\maketitle
\thispagestyle{empty}
\pagestyle{empty}

%%%%%%%%%%%%%%%%%%%%%%%%%%%%%%%%%%%%%%%%%%%%%%%%%%%%%%%%%%%%%%%%%%%%%%%%%%%%%%%%
\begin{abstract}
Deep learning techniques have become prominent in modern fault diagnosis for complex processes. In particular, convolutional neural networks (CNNs) have shown an appealing capacity to deal with multivariate time-series data by converting them into images. However, existing CNN techniques mainly focus on capturing local or multi-scale features from input images. A deep CNN is often required to indirectly extract global features, which are critical to describe the images converted from multivariate dynamical data. This paper proposes a novel local-global CNN (LG-CNN) architecture that directly accounts for both local and global features for fault diagnosis. Specifically, the local features are acquired by traditional local kernels whereas global features are extracted by using 1D tall and fat kernels that span the entire height and width of the image. Both local and global features are then merged for classification using fully-connected layers. The proposed LG-CNN is validated on the benchmark Tennessee Eastman process (TEP) dataset. Comparison with traditional CNN shows that the proposed LG-CNN can greatly improve the fault diagnosis performance without significantly increasing the model complexity. This is attributed to the much wider local receptive field created by the LG-CNN than that by CNN. The proposed LG-CNN architecture can be easily extended to other image processing and computer vision tasks. 
\end{abstract}

%%%%%%%%%%%%%%%%%%%%%%%%%%%%%%%%%%%%%%%%%%%%%%%%%%%%%%%%%%%%%%%%%%%%%%%%%%%%%%%%
\section{Introduction}

Deep learning (DL) has attracted increasing attention for fault detection and diagnosis (FDD) over the last decade. Primarily, the strength of DL lies in its ability to utilize the extensive data present in industrial systems to establish complex models for distinguishing anomalies, diagnosing faults, and forecasting without needing much prior knowledge  \cite{wuest2016machine}. Among various DL methods for FDD, convolutional neural networks (CNNs) have shown great promise due to their efficiency in capturing spatiotemporal correlations and reduced trainable parameters from weight sharing \cite{lecun1995convolutional}. 

Originally developed for image classification, CNNs entail neural networks consisting of convolutions with local kernels and pooling operations to extract features from images \cite{lecun1995convolutional, kiranyaz20211d}. They have also been used in FDD to handle time-series data. Janssens $et$ $al.$ \cite{janssens2016convolutional} made one of the earliest attempts at using CNNs for fault diagnosis. The authors highlighted the capability of CNNs to learn new features from input images converted from time-series data to better classify faults in rotating machinery. Further developments on CNN for FDD can be referred to \cite{huang2019improved, jin2021light, wu2018deep, sun2021gasf, yan2021rotating, luo2020fault}. Note that different from images in computer vision, the images converted from time-series data often possess strong \textit{non-localized features}. To this end, kernels of different sizes, i.e., multi-scale CNN, have been used in \cite{huang2019improved} to cover local receptive fields (LRF) with varying resolutions to improve the learned features. Other techniques such as global average pooling has been employed in  \cite{luo2020fault, nirthika2022pooling} to maintain the integrity of information pertaining to global correlations. However, these approaches either can only directly capture \textit{wider local features} (e.g., multi-scale CNN) or lack learnable parameters in acquiring global correlations (e.g., global average pooling). Thus, they often need to construct deep networks to capture the \textit{global features} that are crucial in multivariate time-series data for FDD \cite{lu2019fault}. In addition, most research studies mentioned above are mainly concerned with 1D or low-dimensional time-series data such as the wheel bearing data \cite{jin2021light}. Research on extending CNN for FDD for high-dimensional multivariate time-series data, e.g., those obtained from chemical processes, still remains limited. One exemplary work is reported in \cite{wu2018deep} where deep CNNs are constructed to diagnose faults from the Tennessee Eastman process (TEP). Gramian angular field is used in \cite{sun2021gasf} to convert multi-dimensional data into multi-channel 2D images to apply CNN for fault diagnosis. Nevertheless, these works still cannot \textit{directly} extract global features from the multivariate time-series data or equivalently, the formed 2D images. Instead, they also rely on constructing deep CNNs to expand the LRF to the entire image for capturing global correlations. As a result, the number of trainable parameters can easily go beyond several millions, causing significant training complexity \cite{wu2018deep,sun2021gasf}. Hence, there is a pressing demand for developing a novel CNN-based FDD framework that adequately extracts global spatiotemporal correlations while maintaining a reasonable number of learnable parameters for multivariate time-series datasets.

This paper proposes a novel local-global CNN (LG-CNN) framework for fault diagnosis for complex dynamical processes. The proposed framework converts multivariate time-series data into images and simultaneously collects both global and local features to classify faults. Local correlations are captured using typical local square kernels, whereas global correlations are integrated using 1D tall (temporal) and fat (spatial) kernels that span the entire height and width of the image. The spatial and temporal global features extracted from the tall and fat kernels are then cohered together to acquire global spatiotemporal patterns in the images. Such global spatiotemporal features are then concatenated with local features extracted with the typical square kernels to merge the information prior to fault diagnosis. The developed LG-CNN is validated with the TEP data and simulation results show that the incorporation of global features into CNN can significantly enhance the diagnosis performance without significantly increasing the model complexity. 

This paper is organized as follows. Section \ref{sec: preliminaries} presents fundamentals about traditional CNNs. The proposed LG-CNN architecture is elaborated in Section \ref{sec: methodology}, followed by a case study of fault diagnosis for TEP in Section \ref{sec: simulation}. The conclusions are given in Section \ref{sec: conclusion}.  
%\textcolor{red}{move them to Section III}In addition, depth-wise convolution with $1\times 1$ kernels have shown to be effective in reduce dimension and computation to allow for exploring deeper and wider architectures \cite{jin2021light, szegedy2015going}. 

\section{Preliminaries}
\label{sec: preliminaries}
In this section, we briefly introduce the main components in a typical CNN including convolutional, pooling, and fully-connected (FC) layers. In addition, batch normalization (BN) is introduced to mitigate the internal covariance shift issues.

\subsection{Convolutional Layers}
\begin{figure*}[tbh]
	\centering
	\includegraphics[width=0.9\textwidth]{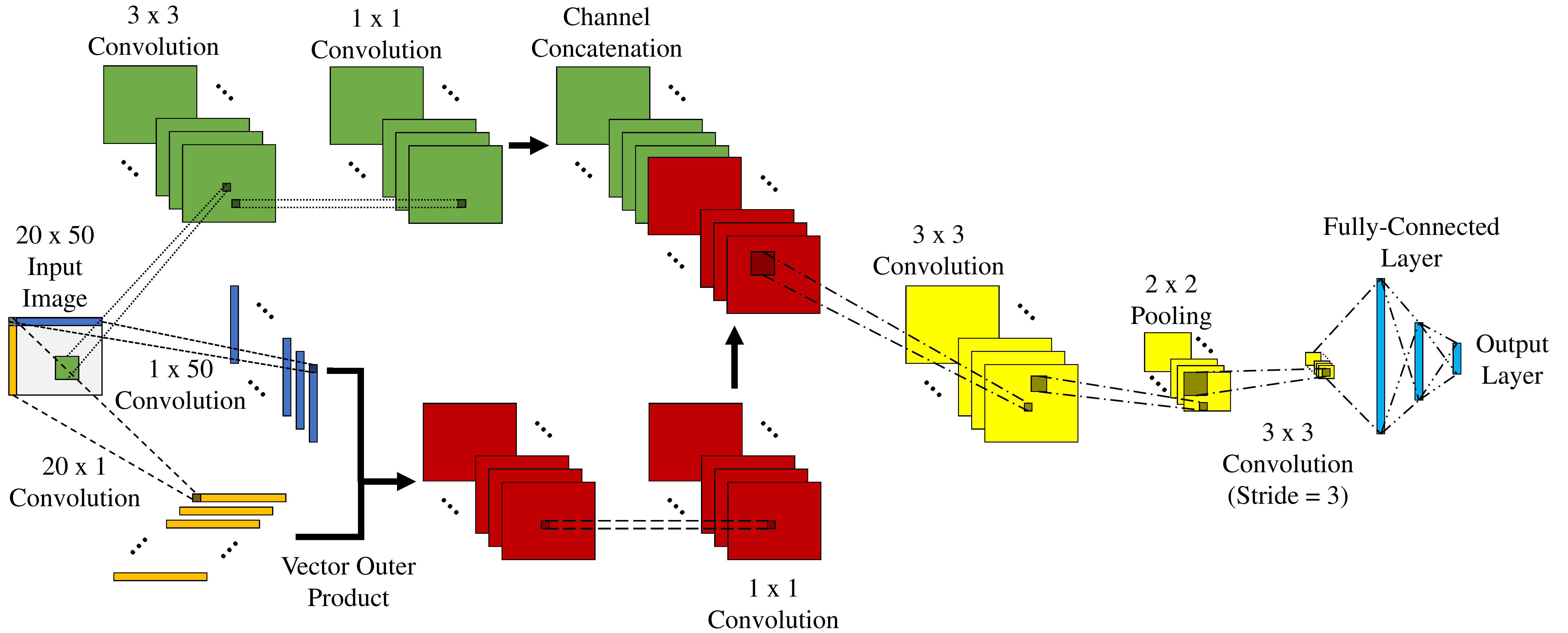}
	\caption{The proposed LG-CNN architecture to extract both local and global features.}
	\vspace*{-5mm}
	\label{fig:VK-CNN}
\end{figure*}

In a convolutional layer, a kernel filter slides across an input feature map where an affine transformation is conducted at every slide location such that: 
\begin{equation}
\mathbf{C}^{l}_{j} = b^{l}_{j} + \sum\nolimits_{i = 1}^{I_{l-1}} \mathbf{X}^{l - 1}_{i} \ast \mathbf{K}^{l}_{i, j}, ~~ j=1, 2, \dots,I_{l},
\label{Conv}
\end{equation}
where $\mathbf{K}^{l}_{i, j} \in \mathbb{R}^{k \times \eta}$ is the kernel of size $k\times \eta$ in layer $l \in  \{1, 2, \dots,  L\}$ and channel $i \in \{1, 2, \dots, I_{l-1}\}$,  $\mathbf{X}^{l - 1}_{i} \in \mathbb{R}^{n \times m}$ is the input feature map of size $n\times m$ to layer $l$. $L$ is the number of layers and $I_l$ is the number of channels in the $l$-th layer. $\textbf{C}^{l}_{j}$ is the $j$-th output map in layer $l$ after the convolution and $b_j^l$ is the bias. The symbol $\ast$ represents the convolution operation. An activation function, such as the rectified linear unit (ReLU), is usually applied to $\textbf{C}^{l}_{j}$ to add non-linearity. Graphically, the first green feature map in Fig. \ref{fig:VK-CNN} illustrates a $3 \times 3$ convolution. The square patch of size $3\times 3$ in the input image represents the LRF of the dark green neuron output in the first feature map in the top branch. Thereby, a LRF can be thought of as the ``field of view" incorporated in calculating a new feature through the convolutional operation. 
\subsection{Pooling Layers}

Pooling operations after convolutional layers often act as a sub-sampling step to reduce dimensionality while preserving information. Specifically, in a localized group of activations on a feature map, pooling summarizes their responses through either averaging or maximizing operations \cite{boureau2010theoretical}. We use max pooling to each local $s \times s$ region of the input feature map $\textbf{X}^{l - 1}_{i}$ $\in$ $\mathbb{R}^{n \times m}$, and the resultant new feature map is shown to be \cite{nirthika2022pooling}
\begin{equation}
\mathbf{P}^{l}_{i} = (\max ~\mathbf{X}^{l - 1}_{i, r})^{S}_{r = 1},
\label{Pool}
\end{equation}
where $\mathbf{P}^{l}_{i}$ is the output of the max pooling operation, and $S$ is the total number of $s \times s$ regions in $\textbf{X}^{l - 1}_{i}$.

\subsection{Fully-Connected Layers}

After all convolutional and pooling layers, the obtained feature maps represent the main features learned by a CNN from an input image. These maps are then flattened into a vector and passed through FC layers for classification or regression. Specifically, the output of the $l$-th FC layer is calculated using
\begin{equation}
a^{l}_{z} = b^{l}_{z} + \textbf{x}^{l - 1} \cdot \boldsymbol{\omega}^{l}_{z},\text{ } z=1, 2, \dots, Z,
\label{Affine}
\end{equation}
where $a^{l}_{z}$ is the output of the $z$-th neuron, $Z$ is total number of neurons in the $l$-th layer, $b^{l}_{z}$ is the bias, $\textbf{x}^{l - 1}$ $\in$ $\mathbb{R}^{\zeta}$ is the activation vector from the previous layer that contains $\zeta$ neurons, $\boldsymbol{\omega}^{l}_{z}\in \mathbb{R}^{\zeta}$ is the weight vector associated with neuron $z$, and $\cdot$ represents the dot product. To add non-linearity, an activation function is applied to $a^{l}_{z}$.    

\subsection{Batch Normalization}

BN accelerates CNN learning by reducing the effects of internal covariance shift \cite{ioffe2015batch}. Layers experience these effects in the learning process when previous layers update their weights and biases resulting in a need to continuously adapt to these changes, and therefore, hinder the learning process. In a 2D-CNN, these effects are mitigated by normalizing the activations from a preceding layer:
\begin{equation}
	\tilde{\textbf{X}}^{l - 1}_{i} = \dfrac{\textbf{X}^{l - 1}_{i} - \mathbb{E}[\textbf{X}^{l - 1}_{i}]}{\sqrt{\mathbb{Var}[\textbf{X}^{l - 1}_{i}]}},
	\label{BN}
\end{equation}
where $\tilde{\textbf{X}}^{l - 1}_{i}$ $\in$ $\mathbb{R}^{n \times m}$ is the $i$-th channel in the $(l-1)$-th layer, $\mathbb{E}[\cdot]$ is the expectation over the training batch and all pixel locations, and $\mathbb{Var}[\cdot]$ is the variance. Then, representation is restored to the layer by the affine computation \cite{ioffe2015batch}
\begin{equation}
\textbf{Y}^{l}_{i} = \tilde{\textbf{X}}^{l - 1}_{i} \alpha^{l}_{i} + \beta^{l}_{i},
\label{BN 2}
\end{equation}
where $\textbf{Y}^{l}_{i}$ $\in$ $\mathbb{R}^{n \times m}$ is the BN output for channel $i$ in layer $l$, and $\alpha^{l}_{i}$ and $\beta^{l}_{i}$ are learnable parameters for each channel.

\section{Methodology}
\label{sec: methodology}
The proposed LG-CNN shown in Fig. \ref{fig:VK-CNN} consists of multi-scale convolutions to extract both local and global features. In addition, we use $1 \times 1$ convolution \cite{jin2021light}, max pooling, and strided convolution \cite{szegedy2015going} for dimension reduction with minimum information loss. 

\subsection{Local Correlations} 
The top branch in LG-CNN (green color in Fig. \ref{fig:VK-CNN}) shows the usage of traditional $3\times 3$ kernels to extract local features from the input image. Note that here we apply BN steps before the ReLU activations. Further, padding is added to ensure that the output has the same dimension as the input image. In addition, to reduce dimensions, $1\times 1$ convolution is conducted to squeeze the number of channels after the convolution. Note that traditional $3\times 3$ kernels usually capture a small LRF region \cite{kiranyaz20211d, yan2021rotating}. The overall mapping from the input image to the extracted feature maps after $1\times 1$ convolutions is abstracted as: 
\begin{equation}
\boldsymbol{\Psi} = f_{\boldsymbol{\theta}_{L}}(\mathbf{X}^{0}) \in \mathbb{R}^{c_{L} \times n \times m},
\label{Br1}
\end{equation}
where $\boldsymbol{\Psi}$ is the feature maps extracted from the top branch, $f_{\theta_{L}}(\cdot)$ represents all operations from input images to the extracted feature maps in the top branch, with a collection of trainable parameters into $\theta_{L}$, and $c_L$ (subscript $L$ represents ``local'') is the channel number in the extracted feature maps. 

\subsection{Global Correlations}
\begin{figure}[tbh]
	\centering
	\includegraphics[width=0.9\columnwidth]{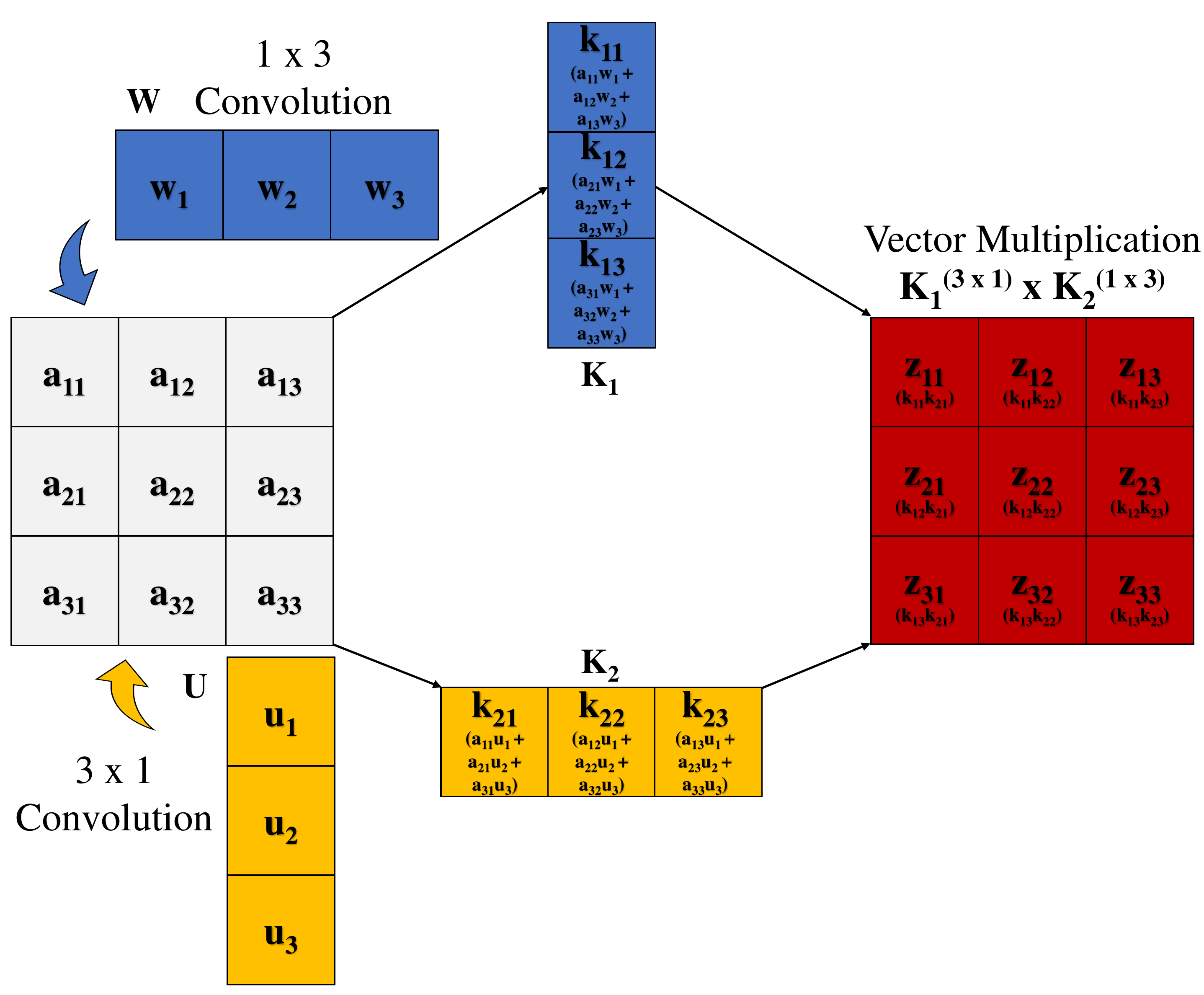}
	\caption{Calculation graph of the 1D tall and fat convolutions, together with the outer product of the resultant feature vectors into 2D maps. Note that the nonlinear activation and batch normalization steps are not included.}
	\vspace*{-5mm}
	\label{fig:Dummy}
\end{figure}

The novelty in the proposed architecture resides in the bottom branch of the LG-CNN model (blue, gold, and red blocks in Fig. \ref{fig:VK-CNN}). To capture global correlations, we propose to use 1D tall (gold) and fat (blue) kernels that encompass the entire width and height of the images so that 
\begin{equation}
\boldsymbol{\Phi} = f_{\boldsymbol{\theta}_{w}}(\textbf{X}^{0}) \in \mathbb{R}^{c_{G}^{\prime} \times n \times 1},~~\boldsymbol{\Omega} = f_{\boldsymbol{\theta}_{h}}(\textbf{X}^{0}) \in \mathbb{R}^{c_{G}^{\prime} \times 1 \times m},	\label{Br2}
\end{equation}
where $\boldsymbol{\Phi}$ and $\boldsymbol{\Omega}$ are the obtained feature maps, $f_{\theta_w}(\cdot)$ and $f_{\theta_h}(\cdot)$ represent the width-wise and height-wise convolutions (see \eqref{Conv}), parameterized by $\theta_w$ and $\theta_h$, respectively, and $c_G^{\prime}$ stands for the channel number after these convolutions (subscript $G$ represents ``global'').  To capture the \textit{global spatiotemporal coherent features}, the acquired feature maps are multiplied together across each channel, to give new features in the form of 2D maps:  
\begin{equation}
\boldsymbol{\Upsilon} = \boldsymbol{\Phi} \otimes \boldsymbol{\Omega} \in \mathbb{R}^{c_{G}^{\prime} \times n \times m}, \label{CBr}
\end{equation}
where $\otimes$ is the outer product of feature vectors obtained after the width-wise and height-wise convolutions, and $\boldsymbol{\Upsilon}$ represents the ultimate feature maps obtained from such multiplication across all $c_G^{\prime}$ channels. For clarity, Fig. \ref{fig:Dummy} presents a calculation graph of the aforementioned operations on a dummy $3 \times 3$ matrix.
Similar to the top branch, $1 \times 1$ convolution takes place after obtaining $\boldsymbol{\Upsilon}$ in \eqref{CBr} to shrink the channel number from $c_{G}^{\prime}$ to $c_G$. In addition, to reduce the risk of divergence or a vanishing gradient due to the multiplication step, BN is applied after it and before the $1 \times 1$ convolution. Inclusive of all the operations detailed, the following description summarizes the overall mapping in the bottom branch:  
\begin{equation}	
\boldsymbol{\Pi} = f_{\boldsymbol{\theta}_{b}}(\textbf{X}^{0}) \in \mathbb{R}^{c_{G} \times n \times m},
\label{EBr}
\end{equation}
where $\boldsymbol{\Pi}$ is the bottom branch's feature maps, $f_{\theta_b}(\cdot)$ entails all the operations in this branch with BN and ReLU implementations, and ${\theta}_{b}$ are all the parameters involved.

\subsection{Max Pooling and Strided Convolution}

After the multi-scale convolutions, feature maps from different branches are concatenated, channel-wise, such that
\begin{equation}
\boldsymbol{\Xi} = [\boldsymbol{\Psi} \text{ } \boldsymbol{\Pi}] \in \mathbb{R}^{(c_{L} + c_{G}) \times n \times m},
\label{Main}
\end{equation}
where $\boldsymbol{\Xi}$ is the combined output from the multi-scale convolutions. Subsequently, a $3\times 3$ convolution layer is conducted to integrate all features, followed by a max pooling step to reduce the dimension. The latter operation employs a $2 \times 2$ pooling with a stride of $2$ to reduce the size of each dimension by half. Interestingly, max pooling does not just serve as a dimensionality reduction step, it also adds non-linearity and regularizes \cite{nirthika2022pooling}. To further reduce dimensionality while at the same time maintaining information, a strided convolution takes place after max pooling. Principally, it is just a typical square convolution with a stride larger than one  \cite{springenberg2014striving}. Here we use a $3 \times 3$ convolution with a stride of 3. As such, the dimension reduction steps can be summarized as
\begin{equation}
\boldsymbol{\Gamma} = f_{\boldsymbol{\theta}_{m}}(\boldsymbol{\Xi}) \in \mathbb{R}^{(c_{L} + c_{G}) \times \kappa \times \epsilon},
\label{Main 2}
\end{equation}      
where $\boldsymbol{\Gamma}$ are the resultant feature maps of dimensions $\kappa \times \epsilon$ with $c_{L} + c_{G}$ channels, which will be flattened and then fed to the classification step, $f_{\theta_m}(\cdot)$ represents the entire mapping parameterized by ${\theta}_{m}$.

\subsection{Fully-Connected Layers}
Lastly, FC layers with one hidden layer concludes the proposed network architecture. To aid with classification, softmax activation is applied to the output vector of the FC layers such that
\begin{equation}
\hat{\boldsymbol{y}} = f_{\boldsymbol{\theta}_{fc}}(vec(\boldsymbol{\Gamma})) \in \mathbb{R}^{C \times 1},
\label{FC}
\end{equation} 
where $\hat{\boldsymbol{y}}$ is the network output after softmax function with $C$ representing the number of classes, $ f_{\boldsymbol{\theta}_{fc}}(\cdot)$ is the overall mapping, and ${\theta}_{fc}$ stacks all learnable parameters in the FC layers. The largest entry in $\hat{\boldsymbol{y}}$ indicates the class that the input image shall be classified into. 

\subsection{Comparison with Related Works}
In the proposed method, the 1D tall and fat kernels are crucial in extracting global features from input images. Using non-square or vector-kernels for enhancing the feature representation has been reported in the literature. For instance, in \cite{chen2020xsepconv}, vector-kernels are employed to expand the LRFs. However, their approach uses local vector-kernels, and thus can only extract local features (the LRF may be wider than square kernels). In our work, the vector-kernels are significantly wider and taller than those in \cite{chen2020xsepconv}. Further, the multiplication step in \eqref{CBr} can allow variables that are far away to be correlated to extract global coherent features from the image. Similarly, in \cite{peng2017large}, vector-kernels are also utilized for augmenting semantic segmentation. Nonetheless, the authors apply 1D fat and tall kernels (but with small size) \textit{in series} to replace traditional square kernels. It shows that their method can expand the LRF but without directly obtaining global patterns. In contrast, our method applies the height-wise tall and width-wise fat kernels \textit{in parallel} and the resultant feature vectors are combined with outer product for acquiring global features. Another related work is reported in \cite{ou2019vector} where traditional square kernels are replaced by tall and fat kernels for popular deep CNN models such as AlexNet and VGG. With such vector-kernels, the number of trainable parameters can be reduced with improved classification performance. However, the work in \cite{ou2019vector} does not study global feature extraction either. 

\section{Simulation}
\label{sec: simulation}
In this section, we use the benchmark TEP data \cite{DVN/6C3JR1_2017} to assess the performance of the proposed LG-CNN. This dataset contains 41 measured and 11 manipulated variables from the simulator of a chemical process consisting of a reactor, condenser, compressor, separator, and stripper \cite{chiang2000fault}. Specifically, the dataset contains 20 different types of faulty data. In this paper, we utilize 47 simulations for each of the 20 faults (40 for training and 7 for testing). The data are sampled every 3 minutes for 25 hours for training and 48 hours for testing. Thus, each training and testing simulation contains 500 samples and 960 samples, respectively. Therefore, a total of 400,000 samples for the training dataset and 134,400 samples for the testing datasets are selected.

\subsection{Preprocessing}
In this stage, variables representing compressor recycle and stripper stream valves are dropped because they remain constant for some simulations \cite{wu2018deep}. In addition, data samples collected before the faults were introduced to each simulation are removed \cite{chiang2000fault}. As a result, only 384,000 training and 112,000 testing samples, with each sample containing 50 variables, are used for training and testing.

To apply the proposed method, the raw multivariate time-series data shall be first converted into images. To do so, every 20 samples from the TEP data are saved as a 2D (gray) image, of dimension $20\times 50$, with one channel. Furthermore, both training and testing images are normalized according to mean and standard deviation of the \textit{training images} (see \eqref{BN}).  Eventually, a training dataset of 19,200 images and a testing dataset of 5,600 image are obtained for this case study.

\subsection{Correlation Map} 
\begin{figure}
	\centering
	\includegraphics[width=0.9\columnwidth]{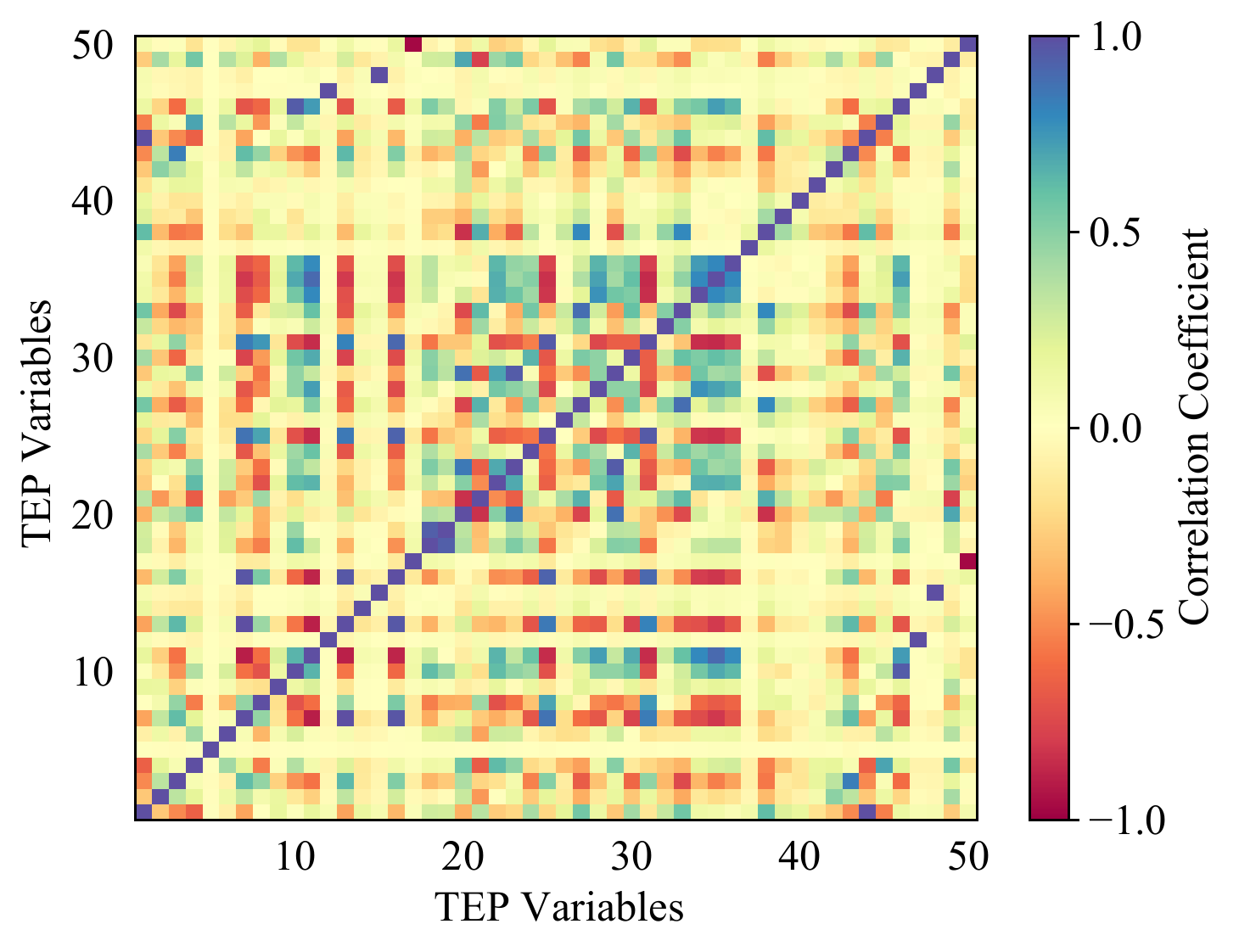}
	\caption{Correlation coefficients of process variables in the TEP.}
	\vspace*{-5mm}
	\label{fig:Corr}
\end{figure}
To study the cross-correlation among process variables, Fig. \ref{fig:Corr} shows the correlation coefficient heat map across all 50 process variables with the data from Fault 2. It is seen that strong correlations are widely scattered, indicating that different columns that are far apart in the obtained images above can possess strong correlations. Thus, both local and global correlations shall be considered when dealing with the images converted from the time-series data.      
 
\subsection{Training and Validation} 

\begin{table}[h]
	\renewcommand\tabularxcolumn[1]{m{#1}}
	\newcolumntype{Y}{>{\centering\arraybackslash}X}
	\vspace*{4mm}
	\caption{Different model structures developed for the LG-CNN.}
	\vspace*{-4mm}
	\label{VK-CNN}
	\begin{center}
		\begin{tabularx}{\columnwidth}{c *{3}{Y}}
			\toprule
			Model & 1 & 2 & 3\\
			\midrule
			& \multicolumn{3}{c}{Input(1, 20, 50)}\\
			\cmidrule(l){1-4}
			Branch 1 & $C$((3, 3), 16)\footnotesize{$^a$} & $C$((3, 3), 32) & $C$((3, 3), 64)\\
			& $C$((1, 1), 8) & $C$((1, 1), 16) & $C$((1, 1), 32)\\
			Branch 2 & $C$((1, 50), 16) & $C$((1, 50), 32) & $C$((1, 50), 64)\\
			Branch 3 & $C$((20, 1), 16) & $C$((20, 1), 32) & $C$((20, 1), 64)\\
			\cmidrule(l){1-4}
			Combined & \multicolumn{3}{c}{Vector Multiplication(Branch 2, Branch 3)}\\
			Branches & & BN & BN\\
			& $C$((1, 1), 8) & $C$((1, 1), 16) & $C$((1, 1), 32)\\
			\cmidrule(l){1-4}
			Main & \multicolumn{3}{c}{Concatenation(Batch 1, Combined Batches)}\\
			& & & $C$((3, 3), 64)\\
			& & & BN\\
			& & $P$((2, 2), 2)\footnotesize{$^b$} & $P$((2, 2), 2)\\
			& & $C$((3, 3), 64) & $C$((3, 3), 128)\\
			& & $s$ = 3\footnotesize{$^c$} & $s$ = 3\\
			&  & BN & BN\\
			& FC(16000, 20)\footnotesize{$^d$} & FC(2304, 300) & FC(4608, 300)\\
			& & FC(300, 20) & FC(300, 20)\\
			\cmidrule(l){1-4}
			Parameters & 321,668 & 719,952 & 1,509,552\\
			\bottomrule
		\end{tabularx}
	\end{center}
	\footnotesize{$^a$ Convolution((Kernel Size), Number of Kernels)}\\ 
	\footnotesize{$^b$ Max Pooling((Kernel Size), Stride)}\\
	\footnotesize{$^c$ Stride}\\
	\footnotesize{$^d$ Fully-Connected(Input Neurons, Output Neurons)}
\end{table}

\begin{table}[h]
	\renewcommand\tabularxcolumn[1]{m{#1}}
	\newcolumntype{Y}{>{\centering\arraybackslash}X}
	\caption{Different model structures developed for CNN.}
	\vspace*{-4mm}
	\label{CNN}
	\begin{center}
		\begin{tabularx}{\columnwidth}{c *{3}{Y}}
			\toprule
			Model & 1 & 2 & 3\\
			\midrule
			& \multicolumn{3}{c}{Input(1, 20, 50)}\\
			\cmidrule(l){2-4}
			& $C$((3, 3), 16) & $C$((3, 3), 32) & $C$((3, 3), 64)\\
			\cmidrule(l){2-4}
			& \multicolumn{3}{c}{BN}\\
			\cmidrule(l){2-4}
			& $C$((1, 1), 8) & $C$((1, 1), 16) & $C$((1, 1), 32)\\
			& & & $C$((3, 3), 64)\\
			& & & BN\\
			& & $P$((2, 2), 2) & $P$((2, 2), 2)\\
			& & $C$((3, 3), 64) & $C$((3, 3), 128)\\
			& & $s$ = 3 & $s$ = 3\\
			&  & BN & BN\\
			& FC(16000, 20) & FC(2304, 300) & FC(4608, 300)\\
			& & FC(300, 20) & FC(300, 20)\\
			\cmidrule(l){1-4}
			Parameters & 320,212 & 716,528 & 1,500,656\\
			\bottomrule
		\end{tabularx}
	\end{center}
	\vspace*{-5mm}
\end{table}

The cross-entropy criterion is used to guide the learning process by minimizing the following objective function:
\begin{equation}
J(\Theta) = \frac{1}{M} \left[\sum_{m=1}^{M} \sum_{c=1}^{C} y_{m} \log P(\hat{y}_{m}=c | x_{m}; \Theta) \right],\\
\label{CEC}
\end{equation} 
where $J$ is the cost, $\Theta = \{{\theta}_{fc}, {\theta}_{m}, {\theta}_{b}, {\theta}_{h}, {\theta}_{w}, {\theta}_{L}\}$ is the collection of all trainable parameters, $M$ is the total number of training images, $C$ is the number of classes, $y_{m}$ is the true label for input $m$, and $P(\hat{y}_{m}=c | x_{m}; \Theta)$ is the predicted probability of classifying the input image $m$ into class $c$ given weights $\Theta$. After training, runs on the testing dataset are used to validate the trained models. For each model structure in Tables \ref{VK-CNN} and \ref{CNN}, 10 training and testing runs are conducted to subside the effects of random variations. All the training and testing are conducted on a NVIDIA RTX A4000 GPU with 64 GB memory. 

\begin{figure}
	\centering
	\includegraphics[width=0.9\columnwidth]{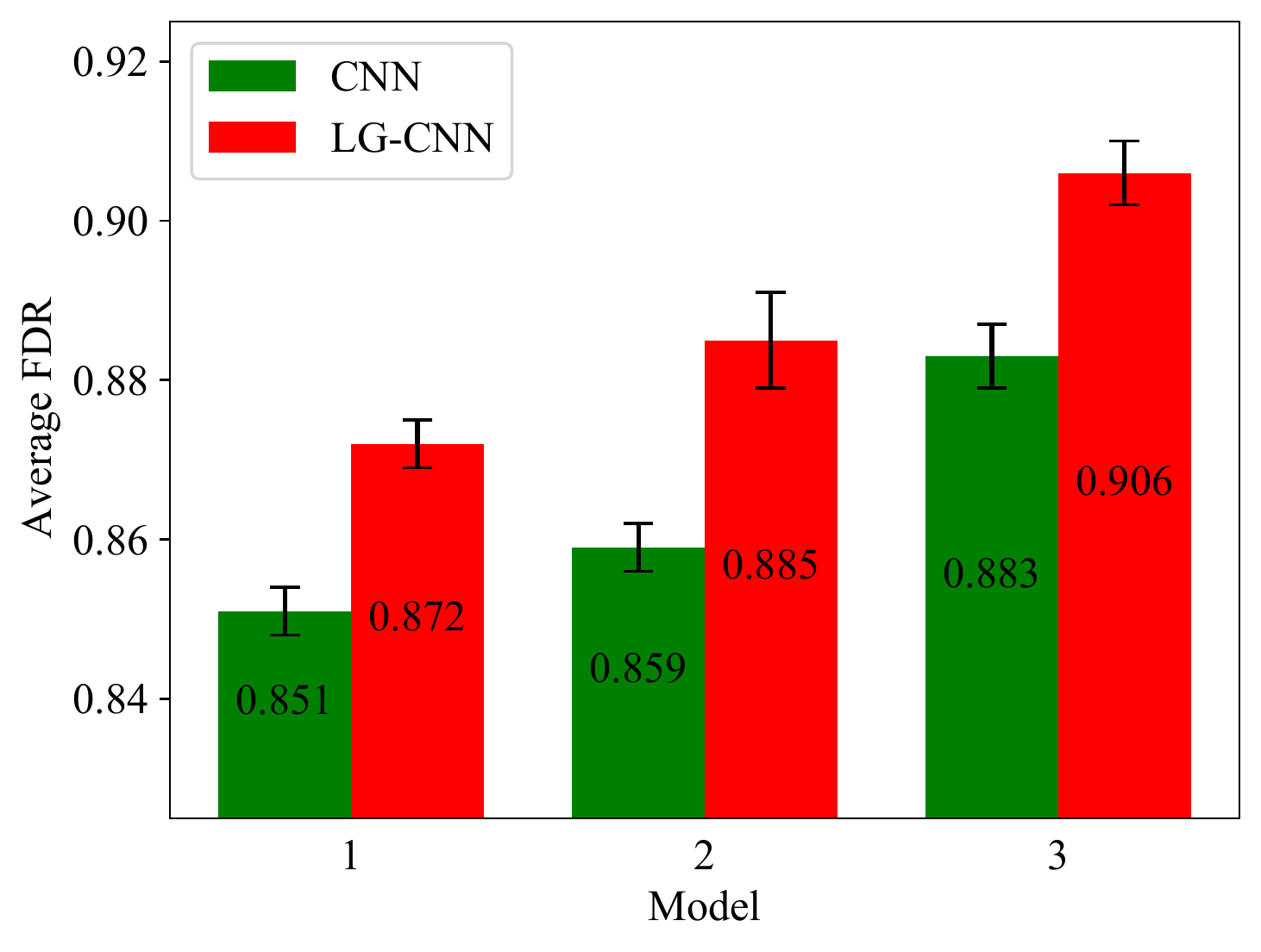}
	\caption{The averaged FDRs from CNN and LG-CNN for the three models.}
	\label{fig:Results}
\end{figure}

\begin{table}[h]
	\renewcommand\tabularxcolumn[1]{m{#1}}
	\newcolumntype{Y}{>{\centering\arraybackslash}X}
	\caption{The FDRs based on LG-CNN and CNN Model 3 for all 20 faults.}
	\vspace*{-4mm}
	\label{CNNvsVKCNN}
	\begin{center}
		\begin{tabularx}{\columnwidth}{c Y c | c Y c}
			\toprule
			Fault ID & CNN & LG-CNN & Fault ID & CNN & LG-CNN\\
			\midrule
			1 & 1.00 & 1.00 & 11 & 0.968 & \textbf{0.975}\\
			2 & 1.00 & 1.00 & 12 & 0.955 & \textbf{0.983}\\
			3 & 0.653 & \textbf{0.760} & 13 & 0.933 & \textbf{0.945}\\
			4 & 1.00 & 1.00 & 14 & 1.00 & 1.00\\
			5 & 1.00 & 1.00 & 15 & 0.475 & \textbf{0.615}\\
			6 & \textbf{1.00} & 0.987 & 16 & 0.791 & \textbf{0.847}\\
			7 & 1.00 & 1.00 & 17 & 0.963 & \textbf{0.966}\\
			8 & 0.949 & \textbf{0.958} & 18 & \textbf{0.927} & 0.915\\
			9 & 0.358 & \textbf{0.416} & 19 & \textbf{0.988} & 0.986\\
			10 & 0.812 & \textbf{0.851} & 20 & 0.886 & \textbf{0.921}\\
			\bottomrule
		\end{tabularx}
	\end{center}
	\vspace*{-5mm}
\end{table}

\subsection{Ablation Experiments}

To assess the fault diagnosis performance, the fault detection ratio (FDR) is defined as the ratio between correct classification and the total sample number for a fault class: 
\begin{equation}
FDR = \dfrac{TP}{TP + FN}, \label{eq: FDR}
\end{equation}
where $TP$ is the number of true positives, and $FN$ is the number of false negatives. For this study, the performance of the proposed LG-CNN, with different model structures as in Table \ref{VK-CNN}, is assessed against that of traditional CNNs of the same structures but without the global feature extraction as in Table \ref{CNN}. Fig. \ref{fig:Results} compares the averaged FDRs from CNN and LG-CNN over all 20 faults for the three developed model structures. It is seen that the LG-CNN clearly outperforms CNN consistently. The only penalty for the LG-CNN is a tiny increase (no more than 0.59\% among all three model structures) in the number of learnable parameters. However, LG-CNN Model 1 gives superior performance (0.872 vs. 0.859) than CNN Model 2, with much less parameters (321,668 vs. 716,528). In addition, LG-CNN Model 2 achieves approximately the same level of FDR (0.885 vs. 0.883) as CNN Model 3, but with almost half of the amount of trainable parameters (719,952 vs. 1,500,656). Thus, with LG-CNN, the model complexity can be significantly reduced to yield comparable performance as CNN, indicating the significance of acquiring those global features. To further compare the performance between LG-CNN and CNN, we use
Table \ref{CNNvsVKCNN} to detail the FDRs for all 20 faults based on Model 3. It is clearly that the LG-CNN model presents better performance across most faults. In particular, the LG-CNN model performs considerably better than CNN for Faults 3, 9 and 15, which are notoriously difficult to diagnose \cite{wu2018deep, sun2021gasf}. 

\begin{figure}
	\centering
	\includegraphics[width=0.9\columnwidth]{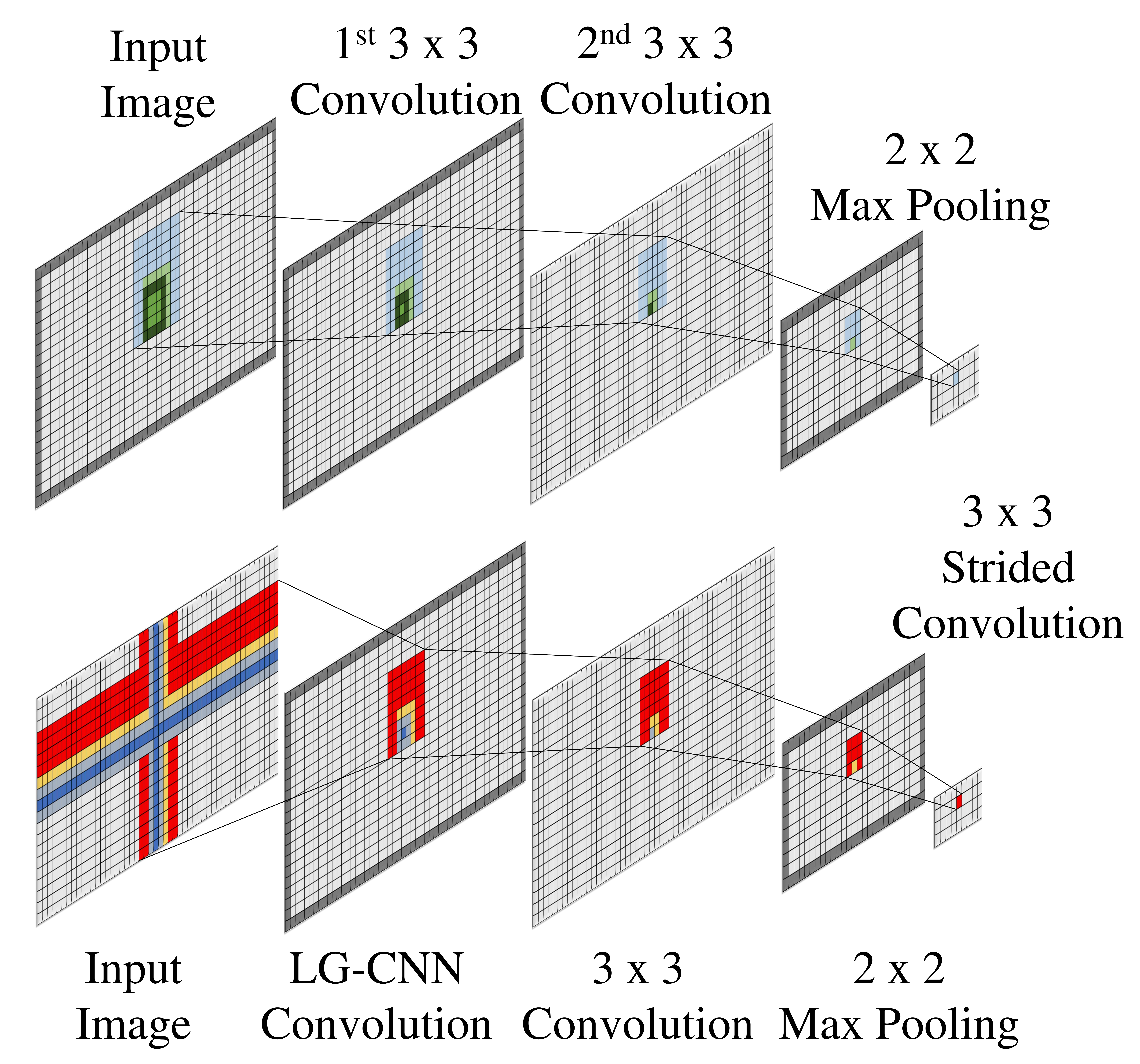}
	\caption{LRFs created by the consecutive convolutional and max pooling operations in CNN (top) and LG-CNN (bottom). The gray boundary around some feature maps represents the paddings. Channels are not included.}
	\vspace*{-5mm}
	\label{fig:LRF}
\end{figure}

All these results clearly indicate the superior performance of our proposed LG-CNN models in fault diagnosis for the TEP. Such observations can be attributed to the large LRFs acquired by employing the 1D height-wise tall and width-wise fat kernels. To further illustrate, Fig. \ref{fig:LRF} demonstrates the LRFs captured by CNN and LG-CNN Models 3. It is observed that the tall and fat kernels from LG-CNN produce an LRF that covers a much wider area in the input image than that from CNN. In other words, the LG-CNN can easily capture global features without building unnecessarily deep networks. Such global features are critical for processing high-dimensional dynamical data that are ubiquitous in industrial processes. 

\section{Conclusion}
\label{sec: conclusion}
This paper proposed a novel LG-CNN model for the fault diagnosis of complex dynamic processes. In the proposed model architecture, local features from the images are captured by traditional square kernels, whereas global features are captured by tall and fat kernels that cover the entire height and width of the image, respectively. Both local and global features are then concatenated in the fully-connected layer for fault diagnosis. The proposed method is validated on a benchmark TEP dataset. Simulation results show that the LG-CNN can significantly improve the performance of fault diagnosis compared with traditional CNN. Moreover, the LG-CNN can employ a much simpler structure than CNN to yield a similar level of fault detection rate as the CNN. This observation lies in the much wider LRF created by LG-CNN than CNN, which is beneficial for obtaining global features. The proposed technique can be easily migrated to other image processing and computer vision tasks due to the simplicity in adding the global feature extraction with vector kernels to traditional CNN. 

\addtolength{\textheight}{-12cm}   % This command serves to balance the column lengths

%%%%%%%%%%%%%%%%%%%%%%%%%%%%%%%%%%%%%%%%%%%%%%%%%%%%%%%%%%%%%%%%%%%%%%%%%%%%%%%%

%%%%%%%%%%%%%%%%%%%%%%%%%%%%%%%%%%%%%%%%%%%%%%%%%%%%%%%%%%%%%%%%%%%%%%%%%%%%%%%%

%%%%%%%%%%%%%%%%%%%%%%%%%%%%%%%%%%%%%%%%%%%%%%%%%%%%%%%%%%%%%%%%%%%%%%%%%%%%%%%%

\section*{Acknowledgment}

S. Al-Wahaibi acknowledges the Distinguished Graduate Student Assistantship and Q. Lu acknowledges the startup funds from the Texas Tech University.

%%%%%%%%%%%%%%%%%%%%%%%%%%%%%%%%%%%%%%%%%%%%%%%%%%%%%%%%%%%%%%%%%%%%%%%%%%%%%%%%

 \bibliographystyle{ieeetr}
 \bibliography{References}

\end{document}